\documentclass{article}
\pdfoutput=1
\usepackage{hyperref}
\hypersetup{bookmarks=false,linkcolor=blue,urlcolor=blue,colorlinks,citecolor=blue}
\usepackage{arxiv}
\usepackage[utf8]{inputenc} % allow utf-8 input
\usepackage[T1]{fontenc}    % use 8-bit T1 fonts
\usepackage{graphicx}
\usepackage[titletoc,title]{appendix}
\usepackage{amsmath}
\usepackage{amsfonts}
\usepackage{mathtools}
\usepackage{scrextend}
\usepackage{array}
\usepackage[table]{xcolor}

\newcommand{\R}{{\mathbb R}}

\makeatletter
\newcommand*{\yncellcolor}{}
\def\yncellcolor\ignorespaces{\@ifnextchar{Y}{\cellcolor{green!25}}{\@ifnextchar{N}{\cellcolor{red!25}}{}}}
\newcolumntype{L}{>{\centering\arraybackslash\yncellcolor}m{2cm}}
\makeatother

\usepackage[
backend=biber,
style=ieee,
natbib=true,
citestyle=numeric
]{biblatex}
\title{Technical Report: \\Auxiliary Tuning and its Application to Conditional Text Generation}
\shorttitle{Technical Report: Auxiliary Tuning and its Application to Conditional Text Generation}

\author{Yoel Zeldes\\
AI21 Labs\\
\texttt{yoelz@ai21.com}
\And
Dan Padnos\\
AI21 Labs\\
\texttt{danp@ai21.com}
\And
Or Sharir\\
AI21 Labs\\
\texttt{ors@ai21.com}
\And
Barak Peleg \\
AI21 Labs\\
\texttt{barakp@ai21.com}
}
\date{Thanks to AI21 Labs leadership for helping guide this report\\[4ex]June 2020}

\begin{document}

\maketitle

\section{TL;DR}

\textbf{What we did:} We designed a simple and efficient method, called Auxiliary Tuning, for adapting a pre-trained Language Model (LM) to a novel task; we demonstrate the approach on the task of conditional text generation. Our approach supplements the original pre-trained model with an auxiliary model that shifts the output distribution according to the target task. 

\textbf{Why it matters:} Achieving state-of-the-art fluency in language tasks such as text generation entails costly training of large LMs~\citep{sharir2020cost}. Auxiliary Tuning allows practitioners to amortize this cost across target tasks by leveraging existing pre-trained LMs. This is done without modifying the pre-trained weights, avoiding the risks of rigidity and catastrophic forgetting, and allowing natural scaling to multiple target tasks.

\textbf{How it works:} The auxiliary model is trained by adding its logits to the pre-trained model logits and maximizing the likelihood of the target task output. Our method imposes no constraints on the auxiliary architecture. In particular, the auxiliary model can ingest additional input relevant to the target task, independently from the pre-trained model's input. Furthermore, mixing the models at the logits level provides a natural probabilistic interpretation of the method. 

\textbf{Results:} Our method achieved similar results to training from scratch for a number of different tasks, while using significantly less compute for training; we share a specific example of text generation conditioned on keywords.

\section{Auxiliary Tuning Framework}

Given a prefix sequence of tokens $x_{<t}$ and the value of some attribute $\alpha$, our goal is to learn a generative model allowing us to sample tokens from the conditional probability $P(x_t | x_{< t}; \alpha)$. The attribute $\alpha$ represents some property that the generated text is expected to possess, such as sentiment, topic or discourse relation. 

Our approach is based on the intuition that learning the conditional probability can be decomposed into two steps:
\begin{itemize}
    \item Learning to generate fluent, natural language; we would like to learn a distribution $P(x_t| x_{< t})$ that assigns high probability to fluent sequences.
    \item Learning to shift the probability distribution $P(x_t| x_{< t})$ as a function of $\alpha$ to obtain $P(x_t | x_{< t}; \alpha)$; we would like the resulting distribution to assign high probability to a subset of fluent sequences that also adhere to the desired attribute.
\end{itemize}

Given a pre-trained LM that accomplishes the first step, we implement the second step by training a model that adds its own predicted logits to those of the pre-trained LM:
\begin{align}\label{eq:1}
P(x_t | x_{< t}; \alpha) &= \mathrm{softmax}\left(\mathrm{logits}_{LM}(x_t | x_{<t}) + \mathrm{logits}_{AUX}(x_t | x_{<t} ; \alpha)\right)
\end{align}

One could combine the \emph{LM} and \emph{AUX} modules earlier in the computation flow (in an earlier hidden layer), and add some layers that learn the interaction. However, logits summation admits a simple interpretation. By applying Bayes' rule we have
\begin{align}\label{eq:2}
P(x_t | x_{< t}; \alpha) &\propto P(\alpha|x_t;x_{<t})P(x_t | x_{<t})
\end{align}

Comparing equations \ref{eq:1} and \ref{eq:2}, we see that the auxiliary module effectively needs to learn the posterior $P(\alpha|x_t;x_{<t})$. Since this task is usually significantly easier to model than the corresponding generative task, we expect to be able to train the auxiliary module with modest data and compute. Moreover, engineering-wise it makes the Auxiliary Tuning framework clean, since each module has a responsibility over a single part of the functionality (single-responsibility principle).

In the next sections we describe how to implement our framework on the conditional text generation task. However, we note that our framework is not limited to text generation. Equation \ref{eq:1} can be generalized to any domain or task, by recognizing that the inputs and outputs can occupy any modality. Therefore our method can be described by the general equation
\begin{align}\label{eq:3}
P(y|x;\alpha) &= g(f_{\textrm{PRE-TRAINED}}(y|x), f_\textrm{AUX}(y|x;\alpha))
\end{align}

where $f_{\textrm{PRE-TRAINED}}$ is the pre-trained model, $f_\textrm{AUX}$ is the auxiliary model, $x$ is the input for the pre-trained model, $\alpha$ is some additional input, $y$ is the output and $g$ is some operation combining the model outputs, which in our case we take to be softmax.

\section{Architecture}

There are no assumptions on the architecture of the auxiliary model, except that it must take $(x_{<t};\alpha)$ as input and output logits in vocabulary space. We chose to use Transformers, similar to the pre-trained LM. Doing so allows us to further benefit from the representation learned in pre-training, i.e., feature extraction - as described below.

\begin{figure}
    \centering
    \includegraphics[width=0.6\linewidth]{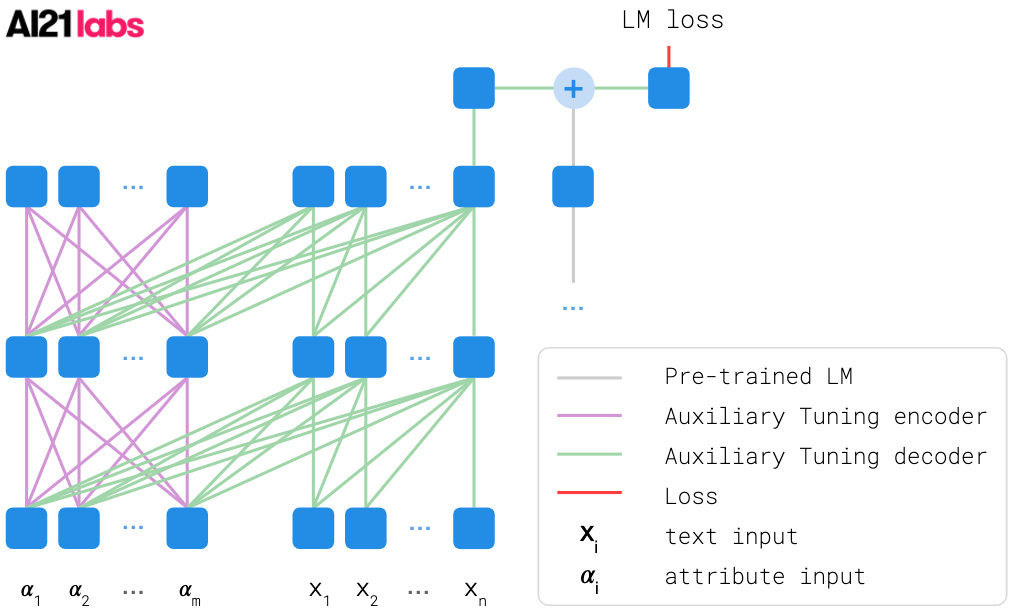}
    \caption{\label{fig:arch}Auxiliary Tuning architecture using Transformers.}
\end{figure}

Let $E_{LM} \in \R^{|V| \times d_{LM}}$ be the embedding table and $T_{LM}$ be the transformers layers
of the pre-trained LM\footnote{We omit positional embeddings for brevity.}, where $|V|$ is the vocabulary size and $d_{LM}$ is the hidden dimension. Similarly,
let $E_{AUX} \in \R^{|V| \times d_{AUX}}$ and $T_{AUX}$ be the token embedding table and transformer layers for the auxiliary model, where $d_{AUX}$ is its hidden dimension, which can be smaller than $d_{LM}$. In the following equations, we treat $x_{<t} \in \R^{(t-1) \times |V|}$ and $\alpha \in \R^{m \times |V|}$ as one-hot representations of the respective sequences\footnote{If the attribute is categorical or numerical, we can simply encode it as a string.}, and define
\begin{align*}
    h^{LM}_{t-1} &= \left(T_{LM}(x_{<t} E_{LM} )\right)_{t-1} \\
    \mathrm{logits}_{LM}(x_t | x_{<t}) &= h^{LM}_{t-1} (E_{LM})^T
\end{align*}
where $h^{LM}_{t-1}$ is the hidden representation of the pre-trained LM, and
\begin{align*}
    h^{AUX}_{t-1} &= \left(T_{AUX}(\left[\alpha E_{AUX} ; x_{<t} E_{AUX} \right])\right)_{m + t-1} \\
    \mathrm{logits}_{AUX}(x_t | x_{<t},\alpha) &= h^{AUX}_{t-1} (E_{AUX})^T
\end{align*}
where $h^{AUX}_{t-1}$ is the hidden representation of the auxiliary model and $[\phantom{a} ; \phantom{a}]$ denotes concatenation along the sequence dimension. Finally, we have
\begin{align}\label{eq:4}
P(x_t | x_{<t}, \alpha) = \mathrm{softmax}(\mathrm{logits}_{AUX}(x_t | x_{<t},\alpha) + \mathrm{logits}_{LM}(x_t | x_{<t}))    
\end{align}
We note that the auxiliary model parameters $E_{AUX}$ and $T_{AUX}$ are trainable whereas $E_{LM}$ and $T_{LM}$ are frozen, i.e., gradients do not propagate to the pre-trained LM. We train the model by maximizing \ref{eq:4}. See illustration of the complete architecture in Figure~\ref{fig:arch}.

\begin{figure}
    \centering
    \includegraphics[width=0.45\linewidth]{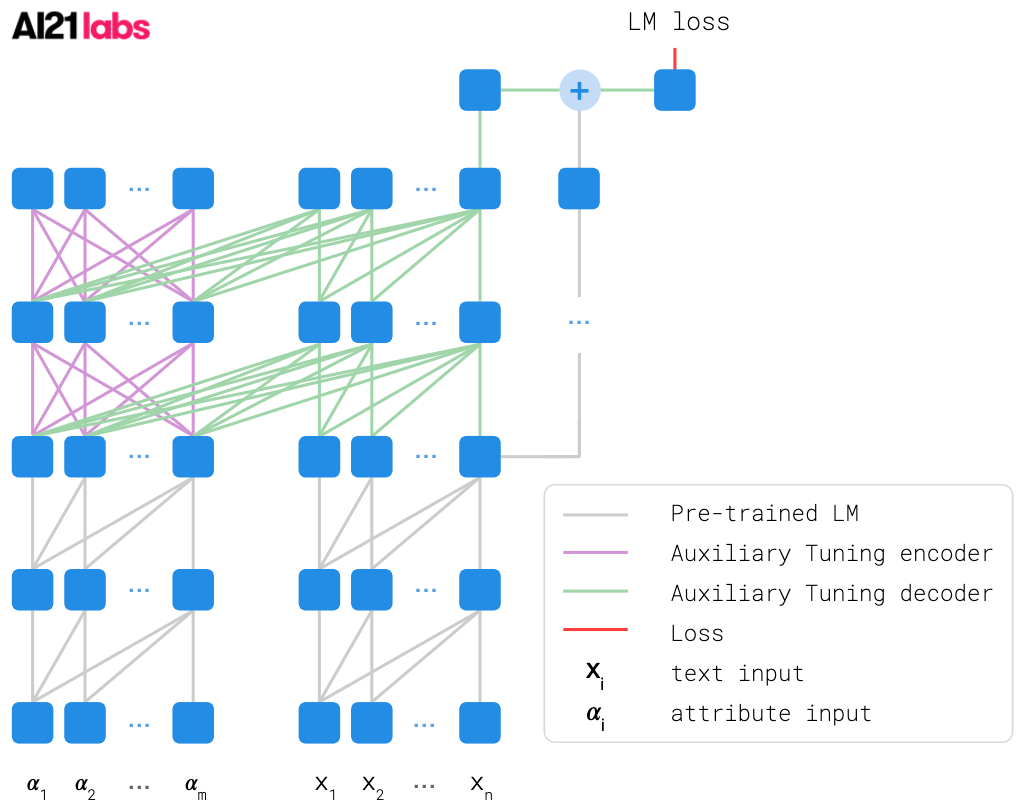}
    \caption{\label{fig:complete_arch}Auxiliary Tuning architecture using Transformers and feature extraction.}
\end{figure}

Next, we show how we can further harness the pre-trained LM as a feature extractor.

Let $Z_{att}$, $Z_{txt}$ be trainable vectors, and $Proj_{AUX \to LM}$ and $Proj_{LM \to AUX}$ be trainable affine projections that map between the hidden spaces of $T_{LM}$, $T_{AUX}$. We denote $T_{LM}^{:L}$ to be the first $L$ layers, where $L$ is a hyper-parameter. Now, we can rewrite $h_{t-1}^{AUX}$ to re-use the lower layers of the pre-trained model for feature extraction
\begin{align*}
h^{att} &= Proj_{LM \to AUX}\left(T_{LM}^{:L}(\alpha E_{LM})\right) + Z_{att} \\
h^{txt}_{<t} &= Proj_{LM \to AUX}\left(T_{LM}^{:L}(x_{<t} E_{LM})\right) + Z_{txt} \\
h_{t-1}^{AUX} &= \left(Proj_{AUX \to LM}\left(T_{AUX}\left(\left[h^{att};h^{txt}_{<t}\right]\right)\right)\right)_{m+t-1} \\
\mathrm{logits}_{AUX}(x_t | x_{<t},\alpha) &= h_{t-1}^{AUX} (E_{LM})^T
\end{align*}

The purpose of the affine projections is to decouple the dimensionality of $T_{AUX}$ from that of $T_{LM}$, resulting in better control of compute and the number of weights that need to be learned. $Z_{att}, Z_{txt}$ allow the model to distinguish between the two inputs (since both have positions that start at zero). The complete architecture is depicted in Figure~\ref{fig:complete_arch}.

\section{Experiments}

We have applied Auxiliary Tuning to the task of text generation conditioned on keywords from a finite inventory. Here is an example input:
\begin{addmargin}[3em]{3em}
    Prefix: \textbf{My salary is really low} \\
    Keyword: \textbf{nevertheless}
\end{addmargin}
Sample output:
\begin{addmargin}[3em]{3em}
    \textbf{My salary is really low} compared to others, \textbf{nevertheless} I find the work incredibly rewarding.
\end{addmargin}

The baseline model is an autoregressive Transformer where keywords are extracted from the training examples and fed as prefix. As a more efficient alternative, we trained an Auxiliary Tuning model on top of our pre-trained language model, \href{https://www.ai21.com/haim-post}{HAIM}, where the keywords are encoded as the additional attribute $\alpha$.

We sampled from the two models and assessed their quality using two metrics:
\begin{itemize}
    \item \textbf{SLOR} \citep{kann-etal-2018-sentence} (a normalized language model score\footnote{We used a LM different from that used by our Auxiliary Tuning model.})~--~a measure of fluency.
    \item \textbf{Accuracy}~--~the fraction of samples that contain the keyword which we conditioned on.
\end{itemize}

\begin{figure}
    \centering
    \includegraphics[width=0.49\linewidth]{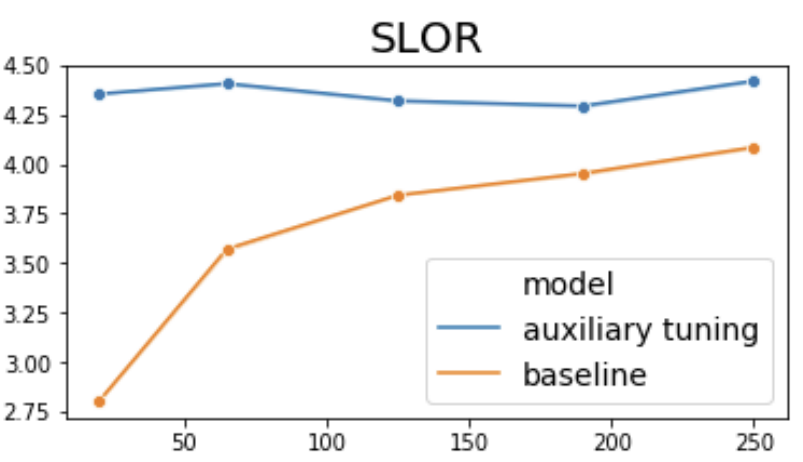}
    \includegraphics[width=0.49\linewidth]{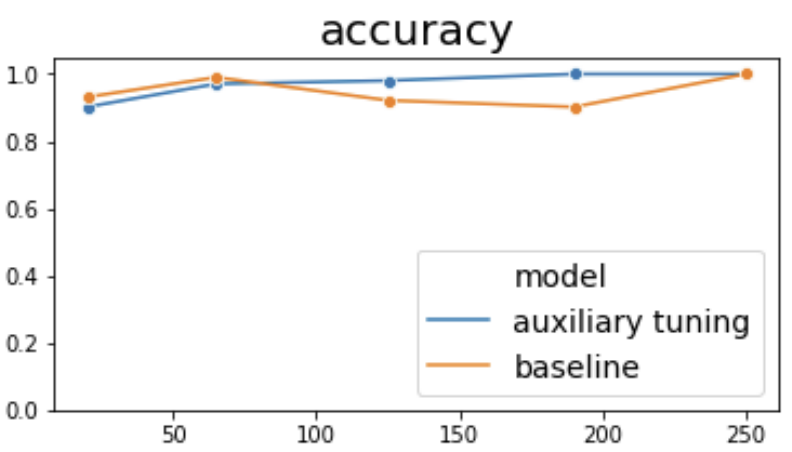}
    \caption{\label{fig:results}SLOR (left) and accuracy (right) of models trained for text generation conditioned on keywords. The x-axis denotes the number of training steps (in thousands).}
\end{figure}

As is seen in Figure~\ref{fig:results}, the models have similar accuracy, but Auxiliary Tuning is much more efficient. As training progresses (indicated by the horizontal axis), the fluency of the baseline improves. However, the auxiliary tuning model generates fluent samples earlier on, demonstrating its effectiveness of harnessing the fluency of the pre-trained LM.

\section{Related Works}

There are a number of existing techniques for adapting pre-trained LMs to novel tasks. We briefly describe them below, and summarize their characteristics in Table~\ref{table:comparison}.

\begin{table}[h]
\begin{center}
\setlength\arrayrulewidth{0.7pt}
\begin{tabular}{|c|L|L|L|L|L|L|}
\hline
& Freezes original LM weights? & Adds trainable parameters? & Allows  feature extraction? & Scalable to multiple tasks? & Scalable model capacity? & Independent task-specific input?\\\hline
\textbf{FS}        & Y & N & N & N & N & N\\\hline
\textbf{FT}        & N & N & N & N & N & Y\footnotemark\\\hline
\textbf{PPLM}      & Y & N & N\footnotemark & Y & N\footnotemark & Y\\\hline
\textbf{ST}        & Y & Y & Y & Y & Y & N\\\hline
\textbf{AT (ours)} & Y & Y & Y & Y & Y & Y\\\hline
\end{tabular}    
\end{center}
\caption{\label{table:comparison}Comparison of approaches for adapting a pre-trained LM to conditional text generation tasks. FS - Few Shot, FT - Fine-Tuning, PPLM - Plug-and-Play Language Model, ST - Side Tuning, AT - Auxiliary Tuning.
} 
\end{table}
\addtocounter{footnote}{-3} %3=n
\stepcounter{footnote}\footnotetext{Fine-tuning supports novel task-specific input if it can be expressed as a text prefix (which is often the case).}
\stepcounter{footnote}\footnotetext{Feature extraction can be used for the discriminative attribute model. Although this should improve attribute relevance, it is unlikely to have any effect on generation quality.}
\stepcounter{footnote}\footnotetext{The discriminative attribute model can be scaled as desired, but this is unlikely to have a major effect on generation quality.}

The de-facto standard approach is to fine-tune the weights of pre-trained LMs on a task-specific dataset. Although fine-tuning has been used successfully in tasks such as domain adaptation~\citep{lee2019patent}, paraphrasing~\citep{witteveen-andrews-2019-paraphrasing} and sentiment control~\citep{ziegler2019finetuning}, it has a number of limitations, as follows.  Fine-tuning modifies the original LM weights. From a machine learning perspective, this is undesirable because it involves navigating a tricky tradeoff between catastrophic forgetting, where the model loses previously learned knowledge (e.g. fluency) and rigidity, where the model is unable to adapt to the target task. These risks are aggravated when scaling fine-tuning to multiple tasks, with the goal of training one model that performs well on all the tasks (under multi-task learning~\citep{ruder2017overview} or incremental learning~\citep{8100070} paradigms). From an engineering perspective, modifying the original LM weights is undesirable because it violates the single-responsibility principle. In addition, common fine-tuning techniques are somewhat limited in providing new inputs to the LM, which is required in controlled text generation.

The Plug-and-Play Language Model (PPLM)~\citep{Dathathri2020Plug} approach aims to adapt a pre-trained LM to conditional text generation without any modification of the model weights and without additional training. Instead, PPLM performs the task adaptation at generation time. They assume access to an additional attribute model $P(\alpha|x)$, a discriminative model that detects whether text $x$ upholds the attribute $\alpha$ corresponding to the control objective, and adjust the hidden state of the LM to increase $P(\alpha|x)$ at each generation step. This technique is relatively complex and involves a number of free parameters that are tuned to trade-off fluency and attribute relevance. 

Side-tuning~\citep{zhang2019sidetuning} also avoids modifying the original model weights. Unlike PPLM, side-tuning adds trainable parameters in the form of a side model that learns a residual on top of the original model. While this approach addresses many of the disadvantages of fine tuning, it still assumes the base and side models are fed exactly the same input. This is a significant limitation for conditional text generation, where the target task adds inputs to express the desired controls.

GPT-3~\citep{GPT3} has shown that task adaptation can be performed on-the-fly in the zero-shot and few-shot settings,  by encoding the conditioning input as a natural language prefix. While the few-shot approach was competitive with supervised baselines on some tasks, it is still unclear to what extent the model is capable of conditioning on novel tasks that do not resemble the pre-training data. In addition, concatenating examples to the model input in the few-shot setting imposes a  limitation on the amount of text that can be generated.

\medskip

\printbibliography

\end{document}